%% file: main.tex
\documentclass[10pt,twocolumn,letterpaper]{article}

\usepackage{cvpr}      


\definecolor{cvprblue}{rgb}{0.21,0.49,0.74}
\usepackage[pagebackref,breaklinks,colorlinks,allcolors=cvprblue]{hyperref}

\usepackage{multirow}
\usepackage[utf8]{inputenc}
\usepackage{tcolorbox}
\usepackage{xcolor}

\tcbset{
    user/.style={
        colback=white, 
        colframe=blue, 
        fonttitle=\bfseries, 
        boxrule=0.5mm, 
        coltitle=blue,
        coltext=black,
        enhanced,
    },
    assistant/.style={
        colback=white,
        colframe=blue,
        fonttitle=\bfseries,
        boxrule=0.5mm,
        coltitle=blue,
        coltext=black,
        enhanced,
    },
}

\def\paperID{5559} 
\def\confName{CVPR}
\def\confYear{2025}

\title{ChatReID: Open-ended Interactive Person Retrieval via Hierarchical Progressive Tuning for Vision Language Models}

\author{Ke Niu\\
Fudan University\\
{\tt\small kniu22@m.fudan.edu.cn}
\and
Haiyang Yu\\
Fudan University\\
{\tt\small hyyu20@fudan.edu.cn}
\and
Mengyang Zhao\\
Fudan University\\
{\tt\small myzhao20@fudan.edu.cn}
\and
Teng Fu\\
Fudan University\\
{\tt\small tfu23@m.fudan.edu.cn}
\and
Siyang Yi\\
Fudan University\\
{\tt\small 22210240353@m.fudan.edu.cn}
\and
Wei Lu\\
Fudan University\\
{\tt\small wlu22@m.fudan.edu.cn}
\and
Bin Li\\
Fudan University\\
{\tt\small libin@fudan.edu.cn}
\and
Xuelin Qian\\
Northwestern Polytechnical University\\
{\tt\small xlqian@nwpu.edu.cn}
\and
Xiangyang Xue\\
Fudan University\\
{\tt\small xyxue@fudan.edu.cn
}
}

\begin{document}
\maketitle
\begin{abstract}
Person re-identification (Re-ID) is a crucial task in computer vision, aiming to recognize individuals across non-overlapping camera views. While recent advanced vision-language models (VLMs) excel in logical reasoning and multi-task generalization, their applications in Re-ID tasks remain limited. They either struggle to perform accurate matching based on identity-relevant features or assist image-dominated branches as auxiliary semantics. 
In this paper, we propose a novel framework \textbf{ChatReID}, that shifts the focus towards a text-side-dominated retrieval paradigm, enabling flexible and interactive re-identification. 
To integrate the reasoning abilities of language models into Re-ID pipelines, 
We first present a large-scale instruction dataset, which contains more than 8 million prompts to promote the model fine-tuning. Next. we introduce a hierarchical progressive tuning strategy, which endows Re-ID ability through three stages of tuning, \textit{i.e.}, from person attribute understanding to fine-grained image retrieval and to multi-modal task reasoning.
Extensive experiments across ten popular benchmarks demonstrate that ChatReID outperforms existing methods, achieving state-of-the-art performance in all Re-ID tasks. More experiments demonstrate that ChatReID not only has the ability to recognize fine-grained details but also to integrate them into a coherent reasoning process.
\end{abstract}

\section{Introduction}
\label{sec:intro}

\begin{figure}[t]
    \centering
    \includegraphics[width=1\linewidth]{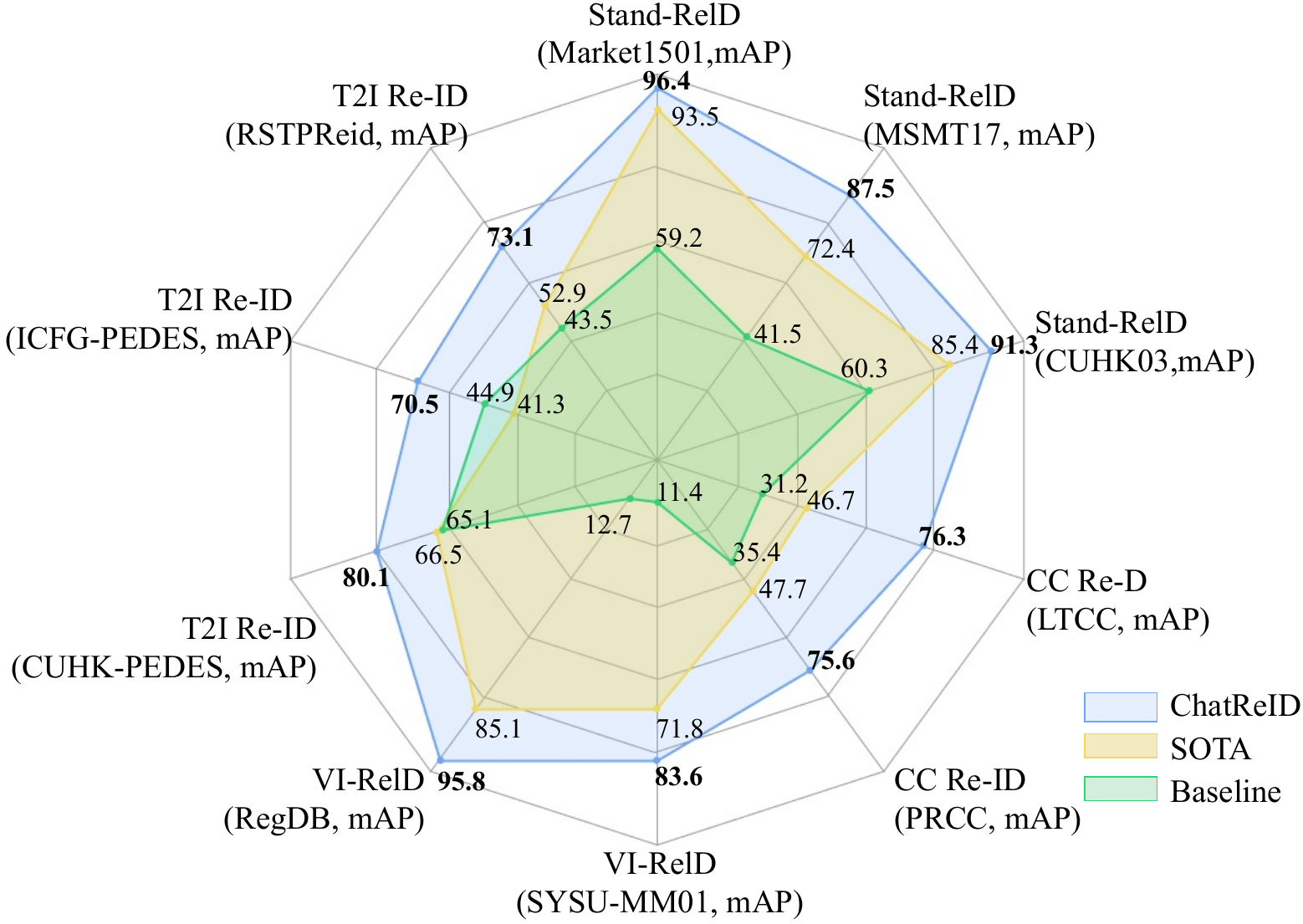}
\caption{
Performance comparisons across ten benchmarks on four person Re-ID tasks show that our ChatReID significantly outperforms previous state-of-the-art methods, demonstrating its superior robustness and effectiveness.
}
\vspace{-0.2in}
\label{performance}
\end{figure}

\begin{figure*}[t]
    \centering
    \includegraphics[width=1\linewidth]{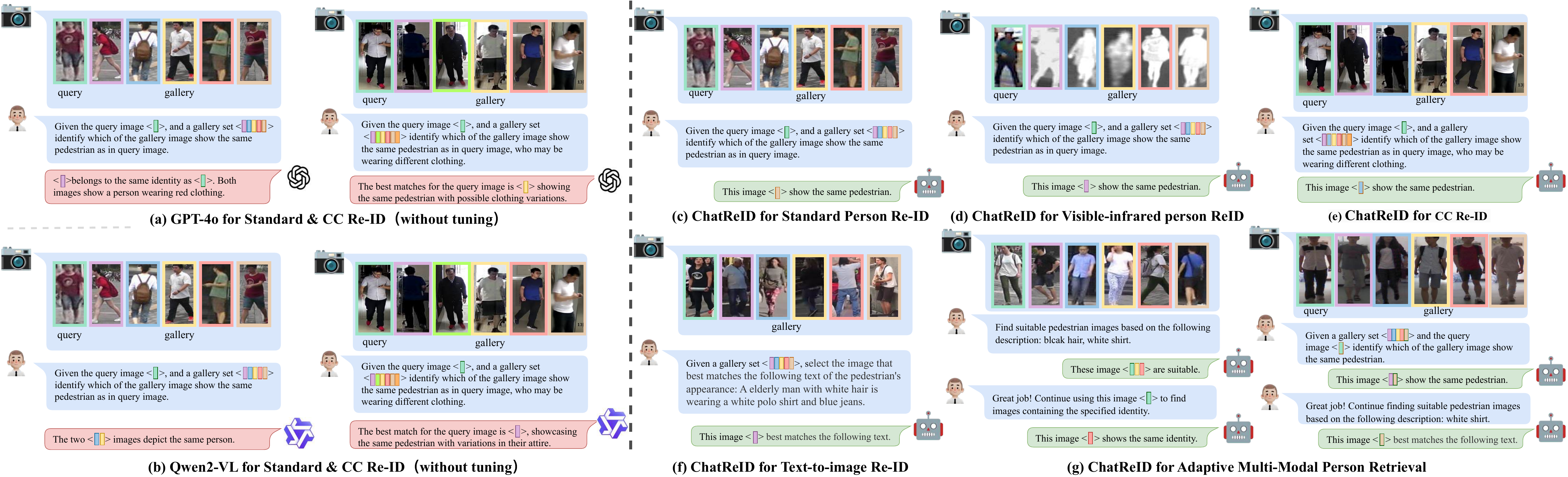}
\caption{(a) and (b) show the results of standard and cloth-changing(CC) Re-ID using GPT-4o and Qwen2-VL without tuning. (c)–(g) illustrate ChatReID’s capabilities across various person Re-ID tasks. Red dialogue boxes indicate incorrect responses, while green dialogue boxes indicate correct responses.~\textbf{Best viewed in color and zoom in.}}
\vspace{-0.2in}
\label{fig:teaser}
\end{figure*}

Person re-identification (Re-ID) is a fundamental task in computer vision that aims to recognize individuals across non-overlapping camera views~\cite{zheng2017person,yuan2020defense,hermans2017defense,niu2025synthesizing}. It plays a crucial role in intelligent surveillance systems, facilitating applications in suspect tracking, crowd management, and access control. In the past decade, person Re-ID has achieved remarkable progress with the development of deep learning techniques~\cite{filax2021influence,zeng2020illumination,fu2023denoising,fu2025foundation}, significantly relieving challenges of pose variants, illumination and occlusions. 

Recently, the emergence of vision-language models (VLMs)~\cite{bai2023qwen,wang2024qwen2,niu2025pht,liu2024visual,yu2025eve,yu2025umit} present new Re-ID paradigms by integrating textual descriptions with visual representations.
They leverage the ability of image-text alignment to enrich feature representations, further promoting the performance and application range of person Re-ID. 
Depending on the usage of text information, VLM-based methods can be broadly grouped into three categories.
(1) Auxiliary supervision using text \cite{li2023clip,yang2024pedestrian}:
they employ text embedding to additionally supervise the feature learning or to integrate with image features.
(2) Text-based person retrieval~\cite{tan2024harnessing,yang2023towards}:
they adopt textual descriptions as queries to retrieve images with the matched identity. 
(3) Text prompts for unified Re-ID~\cite{he2024instruct,he2025instruct}:
they use text prompts or language instructions to manipulate a unified model handling different Re-ID tasks. 
\textit{Despite the impressive effect, these methods still follow the traditional image-side dominated Re-ID paradigm, i.e., extracting features with feature extractors (or encoders) and then sorting them by calculating feature similarity.}

Essentially, person Re-ID is a complex reasoning process that requires analyzing a person's appearance and biological information from an image, then repeatedly comparing different and similar points in a pair of images to determine whether their identities match.
However, existing VLMs, such as Qwen2-VL~\cite{wang2024qwen2} and GPT-4o~\cite{hurst2024gpt}, posses strong vision-language reasoning abilities in programming and communication, yet their potential in person Re-ID tasks remains largely unstudied.
This raises a natural question: \textbf{\textit{can we leverage the advanced reasoning capabilities of VLMs to perform text-side dominated person Re-ID?}}
Intuitively, a straightforward practice is to retain the text decoder of VLMs in the pipeline and ask it to provide the index of the matched images or the similarity score between the given query and gallery images. However, as depicted in Fig.~\ref{fig:teaser}(a) and (b), GPT-4o focuses only on appearance features, leading to incorrect matches, while Qwen2-VL produces almost illogical results. 

We argue that successful person Re-ID requires not only recognizing fine-grained details but also integrating them into a coherent reasoning process. Concretely, 
\textbf{(1)} person identity is inherently an intra-class semantics, meaning that person Re-ID relies on fine-grained biological or identity-related features. While existing VLMs can easily distinguish between broad categories (\textit{i.e.}, cats \textit{v.s.} dogs) or major attributes (\textit{i.e.}, clothing colors), they lack the specialized tuning needed to identify individuals based on subtle visual or textual clues. 
\textbf{(2)} In addition, person Re-ID is an open-set task, where the correct matches can vary depending on the given query or task context. Unlike approaches that use text prompts as conditions, the text-side dominated Re-ID framework requires inferring the desirable similarities through a deep and comprehensive understanding of both textual descriptions and visual content.

In this paper, we explore complex-content reasoning and multi-modal retrieval capabilities of VLMs and introduce \textbf{ChatReID}, a new perspective of using the text decoder for person Re-ID. Our ChatReID is a versatile `one-for-all' framework to interactively ask the model to assist in re-identifying a person given arbitrary, freeform, and necessary descriptions or clues, as shown in Fig.~\ref{fig:teaser}(c)-(g). To overcome the aforementioned challenges, we first propose a hierarchical progressive tuning (HPT) strategy, which consists of three stages. The first stage helps the model understand the attributes or semantics of a person. In the second stage,  the model is guided to learn the ability of image-to-image, image-to-attribute, and text-to-image fine-grained retrieval. The third stage deepens the logical reasoning ability of the model between application scenario descriptions and inputs. Secondly, we create a large-scale instruction dataset with more than 8 million prompts, to promote the tuning of three progressive stages.

\noindent \textbf{Contributions.} We summarize contributions as follows: 
\begin{itemize}
\item We propose ChatReID, a novel framework for person Re-ID that introduces a text-side-dominated retrieval paradigm. ChatReID enables a flexible and interactive retrieval process, enhancing stronger generalization ability.
\item  We present a large-scale instruction dataset and a hierarchical progressive tuning strategy, which endows Re-ID ability through three stages of tuning, \textit{i.e.},  
from person attribute understanding to fine-grained image retrieval and to multi-modal task reasoning.
\item Extensive experiments on ten widely used benchmarks across four different person Re-ID tasks to evaluate the effectiveness of our model. ChatReID achieves state-of-the-art performance in all experiments, outperforming existing methods by a significant margin.
\end{itemize}

\section{Related Work}
\label{sec:formatting}

\subsection{Traditional Person Re-ID}
Person re-identification (Re-ID) is a fundamental task in computer vision, which aims to match the same individual across different camera views based on visual features. Recent studies in person Re-ID carefully designed settings and developed models to tackle every specific scenario.
Standard person Re-ID~\cite{zheng2017person,zheng2017discriminatively,ning2020feature,tan2021incomplete,hermans2017defense,yuan2020defense}, which aims to match individuals across cameras based on visual features. These methods distinguish pedestrian identities based on body posture and appearance.
Cloth-changing Re-ID (CC Re-ID)~\cite{qian2020long,bansal2022cloth,jin2022cloth,hong2021fine,guo2023semantic} is a more challenging variant where individuals change their clothing between camera views. It assists the model in extracting non-clothing information for identity determination. CSSC~\cite{wang2024content} introduces a framework that leverages abundant semantics within pedestrian images to extract identity features. Visible-infrared person ReID (VI-ReID) methods~\cite{feng2019learning,wang2019learning,huang2023deep8vc} extract pedestrian features under low-light environments. DDAG~\cite{ye2020dynamic} improves performance by leveraging intra-modality and cross-modality contextual cues to enhance feature discriminability and robustness to noise. Text-to-image Re-ID~\cite{shao2023unified,han2023text} aims to identify pedestrians based on textual descriptions. It requires the model to understand and align linguistic descriptions with visual attributes. Zhao \textit{et al.}~\cite{zhao2024unifying} proposes a novel method to model multi-modal uncertainty and semantic alignment using Gaussian distributions and a cross-modal circle loss. However, different settings within person Re-ID focus on distinct visual features, making it difficult to effectively integrate these settings into a single model. Consequently, we intend to develop a versatile ‘one-for-all’ framework to interactively ask the machine to help with the person retrieval task.

\subsection{VLM-driven Person Re-ID}
Vision-language models (VLMs)~\cite{bai2023qwen,wang2024qwen2,yang2023dawn,liu2024visual} have garnered significant attention in the AI community due to their impressive generalization capabilities. Recent studies have started investigating the incorporation of VLMs into the person Re-ID paradigm.
Tan \textit{et al.}.~\cite{tan2024harnessing} and Yang \textit{et al.}.~\cite{yang2024mllmreid} primarily focuses on the text-to-image person Re-ID task. The former uses multi-modal large language models (MLLMs) to caption images according to various templates, thereby addressing issues related to the quantity and quality of textual descriptions. The latter proposes a common instruction template and uses features computed by MLLMs to train person Re-ID models. Instruct-ReID~\cite{he2024instruct} is the first work that unifies multiple person Re-ID settings within a single model, generating task-specific instructions and combining instruction encodings with visual encodings for Re-ID training.
Despite significant progress in integrating VLMs into person Re-ID, existing methods face key limitations. Firstly, they fail to fully utilize VLMs' perception and instruction-following abilities. Secondly, many approaches rely on rigid, template-based textual descriptions, limiting adaptability and scalability. Lastly, while some methods unify different Re-ID settings, their flexibility remains constrained, making it difficult to apply them to common scenarios. In this paper, we present a versatile `one-for-all' Re-ID framework that leverages VLMs for interactive, freeform person Re-ID.

\begin{figure*}[t]
    \centering
    \includegraphics[width=1\linewidth]{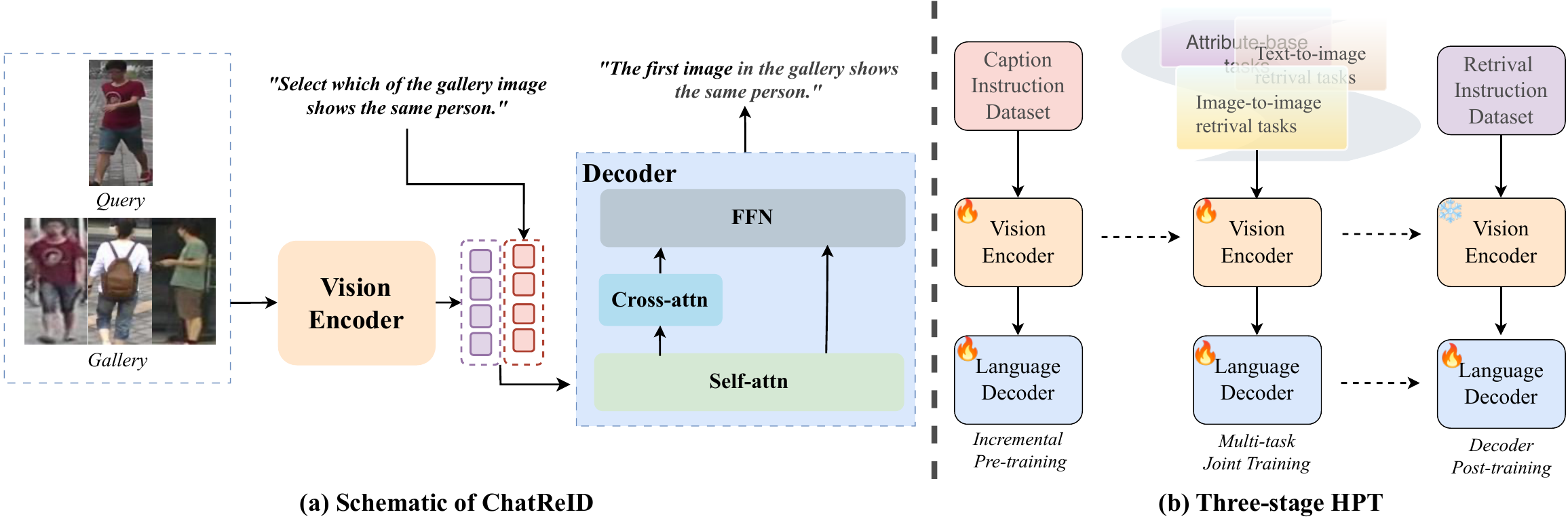}
    \vspace{-0.2in}
\caption{Overview of the ChatReID framework. (a) shows the schematic of ChatReID. (b) shows the three-stage HPT strategy.
}
\vspace{-0.2in}
\label{fig:framework}
\end{figure*}

\section{Methodology}
Our ChatReID is a text-side-dominated framework for person Re-ID. Different from traditional methods that calculates image feature distance as similarity, ChatReID interprets the task requirements from the input text description and performs similarity inference on the given person images accordingly. Finally, it outputs the matched person image in textual form. 

Figure~\ref{fig:framework}(a) illustrates the schematic of our ChatReID, which is primarily composed of a ViT encoder and a LLM decoder. In this section, we start by introducing how to fine-tune the encoder and decoder through our proposed hierarchical progressive tuning strategy (in~\ref{sec:hpt}). Next, we elaborate a large-scale customized instruction dataset built for fine-tuning (see~\ref{sec:data}). Lastly, we discuss the architecture details of our framework, along with the training and inference processes in Sec.~\ref{sec:training}.

\subsection{Hierarchical Progressive Tuning \label{sec:hpt}}

Person Re-ID is a fine-grained retrieval task where identity represents highly abstract semantic information, making it difficult to achieve accurate matches using existing VLMs directly, such as Qwen2-VL~\cite{wang2024qwen2} and GPT-4o~\cite{hurst2024gpt} shown in Fig.~\ref{fig:teaser}(a)-(b).
However, VLMs excel in logical reasoning and multi-task generalization, offering great potential for Re-ID. To bridge this gap, as shown in Fig.~\ref{fig:framework}(b), we introduce a hierarchical progressive fine-tuning strategy, which gradually enhances the model’s Re-ID capability through three stages: (1) person attribute understanding, (2) fine-grained image retrieval, and (3) multi-modal task reasoning.

\subsubsection{Stage One: Person Attribtue Understanding}

Considering the diversity of person images, the first stage focuses on improving the model's ability to understand pedestrians by learning and extracting fine-grained attributes such as gender, clothing, and posture.

\noindent \textbf{$\bullet$ Image captioning.} Following previous studies~\cite{liu2023visual,li2022blip}, we
use image captioning as a foundational fine-tuning task. Specifically, the model is trained to generate detailed attribute descriptions based on a given person image and corresponding prompts. To further guide the model, we empirically incorporate several predefined key attributes into the prompt. This design not only encourages the model to capture more precise and informative attribute descriptions but also potentially strengthens its ability to differentiate identities. The prompt used is:

\textit{``In the \{person image\} provided, can you give a detailed description of the pedestrian, including their gender, age range, hair, type and color of clothing, footwear type and color, posture or gait, any patterns or accessories, and any distinguishing features?'' }

\subsubsection{Stage Two: Fine-grained Image Retrieval}

After completing the first stage of tuning, we obtain a vision encoder that can effectively capture fine-grained person attributes. To further promote its ability for person re-identification, we introduce several fine-tuning tasks in the second stage to guide the model in matching identity-related features with fine-grained information. Specifically, we structure tasks into three categories of retrieval tasks: \textit{image-to-image}, \textit{text-to-text}, and \textit{image-to-attribute}.

\begin{figure*}[t]
    \centering
    \includegraphics[width=1\linewidth]{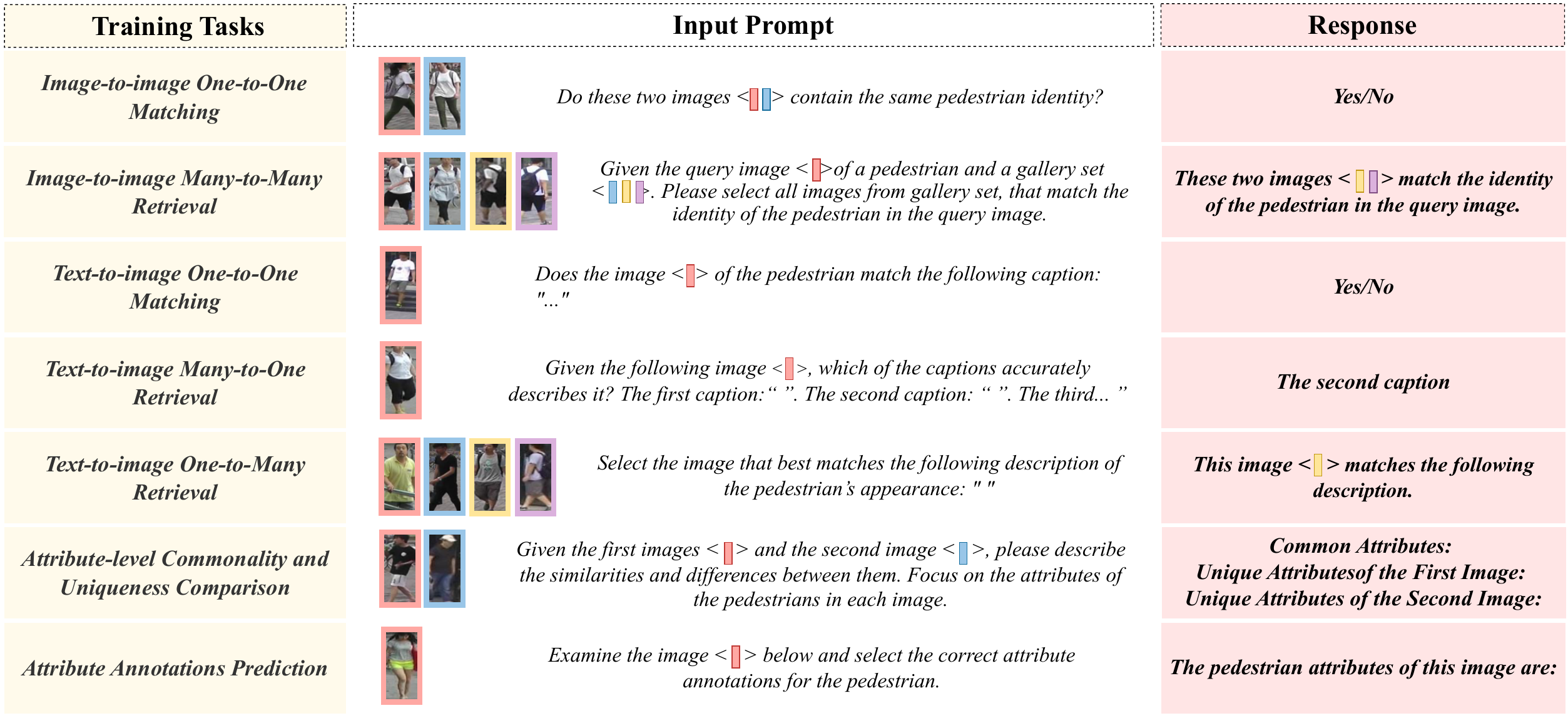}
\caption{Multi-task Design Diagram for joint training in the stage 2. Concretely, we conduct seven distinct matching and retrieval tasks between text and image modalities, encouraging  VLMS to acquire an initial capability for fine-grained image retrieval based on images, textual descriptions, and pedestrian attributes.
}
\vspace{-0.2in}
\label{fig:multi}
\end{figure*}

Additionally, each type of retrieval task is designed with varying levels of complexity, including \textit{one-to-one}, \textit{one-to-many}, and \textit{many-to-many} scenarios. This diverse range of task complexities is crucial for balancing training difficulty and improving the model’s robustness (see our discussions in Sec.~\ref{sec:ablation}). Figure~\ref{fig:multi} illustrates seven fine-tuning tasks, including examples of prompts and responses, involved in the second stage.

\noindent \textbf{$\bullet$ Image-to-image Retrieval.} This task has two levels of \textit{one-to-one} and \textit{one-to-many}. For the one-to-one level, we sample two images and ask the model to determine whether the people in these two images are the same person. To balance the training, we set a probability of 0.5 to sample images with positive answers. For the one-to-many level, we randomly select one image as the query and choose a set of $N$ images, where $N$ images include $n$ images that have the same identity. The model is required to identify all matched images. During training, values of both $N$ and $n$ vary to make the task more challenging and diverse.

\noindent \textbf{$\bullet$ Text-to-image Retrieval.} This task aims to identify the matched person image based on a text description, which has three levels of \textit{one-to-one}, \textit{many-to-one}, and \textit{one-to-many}. Similar to the image-to-image retrieval task, the one-to-one level requires the model to determine if the image and the text description match or not (`yes' or `no'). And for the last two levels, both need to find the correctest answer from a set of options.

\noindent \textbf{$\bullet$ Image-to-attribute Retrieval.}
This task is proposed to learn to distinguish individuals based on specific attributes, and we design two specific levels. For the first one, two images are sampled and the model is tasked with explaining the similarities and distinctions in attributes of two pedestrians. It is expected to further enhance the model's understanding ability of person attributes.
For the second, an image is randomly selected and fed into the model to generate its attribute descriptions or annotations (\textit{e.g.}, gender, clothing color, etc.). It is similar to the first stage, with the intention to deepen and consolidate this capability of the model in the current stage.

\subsubsection{Stage Three: Multi-modal Task Reasoning}
Stage Three enhances ChatReID’s rasoning capability of Re-ID objectives and improves its instruction-following capabilities for practical applications. 
After compressing the fine-grained pedestrian discrimination information via the above two stages, ChatReID has foundational skills in person identity matching.
In this stage, we directly adopt the training objective of 4 person Re-ID tasks.

\noindent \textbf{$\bullet$ Standard person Re-ID.} Standard person Re-ID is a image-to-image retrieval task, where a query image and a set of gallery images are provided, and the goal is to identify the pedestrian in the gallery that matches the identity of the query, the prompt used is:

\textit{``Given the \{person image\} provided, identify which of the following images show the same pedestrian as in the first image.''}

\noindent \textbf{$\bullet$ CC person Re-ID.} Based on the cloth-changing person Re-ID setting, the prompt used is:

\textit{``Given the \{person image\} provided,  select which of the \{gallery\} shows the same person, who may be
wearing different clothing.''}

\noindent \textbf{$\bullet$ VI person Re-ID.} Based on the VI person Re-ID setting, the prompt used is:

\textit{``Using the given \{person image\}, choose the corresponding \{gallery\} that matches the same pedestrian, considering both infrared and visible light images for re-identification.''}

\noindent \textbf{$\bullet$ T2I person Re-ID.} T2I person Re-ID is a text-to-image retrieval task, where a pedestrian image description is given, and the goal is to find the matching pedestrian image in the gallery based on the description, the prompt used is:

\textit{``Select the \{person image\} that best matches the following description of the pedestrian's appearance: \{text\} \.''}

To improve robustness and prevent bias from fixed-length galleries, we introduce variability in the gallery length, randomly adjusting it while ensuring at least one image matches the query identity.

\subsection{Data Engine \label{sec:data}}
For each training stage, we collect and organize a substantial number of person Re-ID datasets to construct the instruction tuning datasets used for training. We utilize a total of 19 open-source person Re-ID datasets. 
The complete list of datasets and corresponding statistical information can be found in the supplementary materials.

In Stage 1, we utilize over 5M image-text instruction pairs. 
To address the low-quality problem, we performed post-processing to reduce noise and enhance data quality. The data construction process involved several key steps:
\textbf{Image Filtering by Quality}. We filter images based on size and resolution to address blurriness and improve overall quality.
\textbf{Image Filtering by Attributes}. We employ GPT-4o~\cite{hurst2024gpt} to predict pedestrian attributes. If the model failed to recognize the pedestrian’s clothing. These are marked as occluded and removed from pre-training.
\textbf{Image Captioning}. Using GPT-4o~\cite{hurst2024gpt}, we generated high-quality, diverse textual descriptions for each image.

In Stage 2, we use seven datasets from various Re-ID settings. We constructed a joint training instruction dataset comprising 3M samples.
Given that the current LLM model demonstrates superior text-to-image retrieval capabilities compared to image-to-image retrieval in the person Re-ID task, we augment the data volume for the image-to-image tasks. The two image-to-image tasks account for approximately 25\% to 30\% of the overall data volume, while the remaining tasks each contribute roughly 10\%. 

\noindent\textbf{Gallery image sampling}
The quantity of sampled gallery images for the Many-to-Many Retrieval task adheres to a uniform distribution $N \sim \text{Uniform}(a, b)$. 
The distribution of the response $P(n)$ is obtained based on inverse transform sampling. This approach allows us to balance the probability that each number falls within different intervals, ensuring that the probability of recording any number eventually is uniform across all numbers. The probability \( P(n = k) \) is given by:
\begin{equation} 
P(n = k) = \frac{1}{k - 1}, \quad \text{for } k \in {2, 3, \dots, N} 
\end{equation}

In Stage 3, we gather ten widely used benchmarks across four Re-ID settings to create a comprehensive instruction dataset with 500K samples and three specific training objectives. During the training phase, we utilized only the training subsets of these benchmarks, while the inference phase involved evaluation using their respective test sets.

\subsection{Details of training and testing \label{sec:training}}
In this section, we detail the technical architecture of ChatReID. The ChatReID framework, illustrated in Figure~\ref{fig:framework}, integrates the structure of Qwen2-VL~\cite{wang2024qwen2} due to its Naive Dynamic Resolution architecture, which effectively addresses the challenges posed by variable image resolutions in person Re-ID tasks.

In Stage 1, both the encoder and decoder are initialized with a pre-trained VLM to leverage existing language-vision knowledge. We conduct full-parameter training on both the vision encoder and language decoder of Qwen2-VL 2B to enhance the encoder’s focus on pedestrian features, while training the decoder ensures that captions accurately reflect pedestrian attributes.
We standardize images with resolutions below 256$\times$128, resizing them to 256$\times$128 while leaving higher-resolution images unchanged. This adjustment leverages our experience with pedestrian tasks, and moderate image compression enhances training efficiency.

In Stage 2, we carry over the encoder from Stage 1 but replace the decoder to continue full-parameter training. The previous decoder’s strong captioning focus could hinder outcomes in this stage, so we use a freshly pre-trained decoder instead. During this stage, we adjust the size of input images to accommodate a large number of gallery images. Images larger than 384$\times$192 were rescaled to 384$\times$192, while images below this threshold were left unaltered.

In Stage 3, we inherit the full structure from Stage 2, training with the encoder frozen. With pedestrian recognition skills already developed, this stage focuses on enhancing the model's instruction-following capabilities. The input image resolution is preserved without any adjustments.

We rethink the assessment of the person Re-ID problem and formulate a question construction in the style of Visual Question Answering (VQA) utilizing a multi-modal language model from a practical application perspective. The instruction tuning data amounts to 500K.

\noindent\textbf{Person Re-ID.} The most common type of query involves providing a text or image query alongside a gallery set, with the goal of identifying images in the gallery that match the query’s identity. However, for current VLMs, retrieving an indefinite-length sequence of images from an indefinite gallery set based on specific criteria is highly challenging. To address this, we transform this complex retrieval task into a multi-turn best-choice problem through engineering enhancements. Specifically, at each step, the model selects the most similar image, making a locally optimal choice, which simplifies the problem and reduces computational complexity.

\noindent\textbf{Attribute-Based Person Retrieval.} In real-world applications, providing a fully detailed and precise query, whether text or image, is often difficult and costly. Users typically possess only fragmented textual information, such as "a person wearing a blue top." To accommodate this limitation, we developed a retrieval objective based on specific pedestrian attributes, enabling effective retrieval even when the query contains only partial details. Given a gallery set and a query with accurate pedestrian attribute descriptions, the model identifies the gallery image that best matches the query information.

\noindent\textbf{Adaptive Multi-Modal Person Retrieval.} Finally, in practical scenarios, multi-turn retrieval based on limited query information often results in a set of responses with relatively low accuracy. As retrieval continues, users typically gather more query details to improve subsequent searches. For example, beginning with attribute-based retrieval, if the correct match appears among the initial results, the user can use that match as a new query to refine further results. Alternatively, users can start with image-based retrieval and add textual details to address complexities such as clothing changes or long-term recognition tasks. By integrating multi-modal information, this approach significantly improves both the flexibility and accuracy of person Re-ID, advancing its real-world applicability. To support this, we design a training objective for handling mixed queries with multi-modal inputs.

\section{Experiments}

\subsection{Experimental Setup} Implementation details are provided in the supplementary.

\noindent \textbf{Datasets}. We use the test sets from the datasets employed in Stage 3. The evaluation is conducted using standard metrics in person Re-ID, including Cumulative Match Characteristic (CMC) and Mean Average Precision (mAP). Results are reported in terms of mAP and Rank-1 accuracy.

\noindent \textbf{Evaluation strategy}.
Evaluating VLMs for person Re-ID in a VQA format presents two primary challenges. First, the large gallery sets in person Re-ID datasets often exceed the token limits of current VLMs. Second, retrieving a sequence of images of variable length based on specific criteria from such vast galleries is inherently difficult for these models.
To address these challenges, we implement several engineering optimizations. We first adopt a baseline model (\textit{e.g.}, ResNet-50) to filter the gallery by calculating feature similarity between the query image and gallery images. Only images with similarity above a threshold $\tau$ are retained, significantly reducing the gallery size while maintaining accuracy and improving evaluation efficiency. Next, we reformulate the retrieval task as a multi-turn best-choice problem. Finally, we sample non-overlapping images from the filtered gallery, pairing each selected image with the query and ranking them based on response confidence, ultimately producing the final similarity list.

\begin{table}[t]
    \centering
    \small
    \caption{Comparison of ChatReID with SOTA methods across three standard person Re-ID datasets.}
    \vspace{-0.1in}
    \label{stand}
    \setlength{\tabcolsep}{0.8mm}{
    \begin{tabular}{lcccccc}
        \toprule
        \multicolumn{1}{c}{\multirow{2}[0]{*}{\textbf{\textsc{Methods}}}} & \multicolumn{2}{c}{\textbf{Market1501}} & \multicolumn{2}{c}{\textbf{MSMT17}} & \multicolumn{2}{c}{\textbf{CUHK03}} \\
         \cmidrule(r){2-3} \cmidrule(r){4-5} \cmidrule(r){6-7}
                 & mAP & Rank-1 & mAP & Rank-1 & mAP & Rank-1 \\
        \midrule
        TransReID~\cite{he2021transreid}   & 86.8 & 94.4 & 61.0 & 81.8 & - & - \\
        SAN~\cite{jin2020semantics}         & 88.0 & 96.1 & - & - & 76.4 & 80.1 \\
        HumanBench~\cite{tang2023humanbench}  & 89.5 & - & 69.1 & - & 77.7 & - \\
        PASS~\cite{zhu2022pass}        & 93.0 & 96.8 & 71.8 & 88.2 & - & - \\
        IRM~\cite{he2024instruct}         & 93.5 & 96.5 & 72.4 & 86.9 & 85.4 & 86.5 \\
        \midrule
        ChatReID    & \textbf{96.4} & \textbf{97.2} & \textbf{87.5} & \textbf{90.1} & \textbf{91.3} & \textbf{92.7} \\
        \bottomrule
    \end{tabular}}
    \vspace{-0.1in}
\end{table}

\begin{table}[t]
\centering
\small
\caption{Comparison of ChatReID with SOTA methods across two cloth-changing Re-ID (CC Re-ID) datasets.}
\vspace{-0.1in}
\label{cc}
\setlength{\tabcolsep}{3mm}{
\begin{tabular}{lcccc}
\toprule
\multicolumn{1}{c}{\multirow{2}{*}{\textbf{\textsc{Methods}}}} & \multicolumn{2}{c}{\textbf{LTCC}} & \multicolumn{2}{c}{\textbf{PRCC}}  \\ 
\cmidrule(r){2-3} \cmidrule(r){4-5}
                         & mAP & Rank-1 & mAP & Rank-1  \\ \midrule
HACNN~\cite{li2018harmonious}          & 26.7 & 60.2 & - & 21.8 \\ 
RGA-SC~\cite{zhang2020relation}         & 27.5 & 65.0 & - & 42.3  \\ 
PCB~\cite{sun2018beyond}            & 30.6 & 65.1 & 38.7 & 41.8 \\ 
IANet~\cite{hou2019interaction}          & 31.0 & 63.7 & 45.9 & 46.3  \\ 
CAL~\cite{gu2022clothes}             & 40.8 & 74.2 & - & -  \\ 
TransReID~\cite{he2021transreid}      & - & - & 44.2 & - \\
IRM~\cite{he2024instruct}                & 52.0 & 75.8 & 52.3 & 54.2 \\ \midrule
ChatReID  & \textbf{76.3} &\textbf{82.7}   &\textbf{75.6} &\textbf{80.2} \\
  \bottomrule
\end{tabular}}
\vspace{-0.1in}
\end{table}

\subsection{Experimental Results}

\textbf{Standard Person Re-ID.}
As shown in Tab.~\ref{stand}, ChatReID achieves remarkable results across all standard person Re-ID datasets. The most notable improvement is observed on the challenging and large-scale MSMT17 dataset, where ChatReID achieves a 15.1\% improvement in mAP. This substantial gain demonstrates the effectiveness of our three-stage tuning strategy, which effectively extracts robust, pedestrian-specific features, thereby improving the model’s resilience in demanding scenarios. Such performance improvement further validates the effectiveness of ChatReID.

\noindent \textbf{Cloth-changing Re-ID (CC Re-ID).}
The results in Tab.~\ref{cc} indicate that ChatReID demonstrates significant mAP enhancements over IRM on both the LTCC and PRCC datasets, with increases of 24.3\% and 23.3\% in mAP, respectively. 
Based on our analysis, this improvement can be attributed to two main reasons.
First, all three tuning stages are based on image-text pairs, where textual information aids the visual encoder learn variations in clothing more effectively. Second, we introduced innovative attribute-based training tasks in Stage 2, which enhance the model's ability to recognize CC scenarios.

\begin{table}[t]
\small
\centering
\caption{Comparison of ChatReID with SOTA methods across two visible-infrared person ReID (VI-ReID) datasets.}
\vspace{-0.1in}
\label{vi}
 \setlength{\tabcolsep}{3.2mm}{
\begin{tabular}{lcccc}
\toprule
\multicolumn{1}{c}{\multirow{2}{*}{\textsc{Methods}}} & \multicolumn{2}{c}{\textbf{SYSU-MM01}} & \multicolumn{2}{c}{\textbf{RegDB}}  \\
\cmidrule(r){2-3} \cmidrule(r){4-5}
                         & mAP & Rank-1 & mAP & Rank-1  \\ \midrule
DART~\cite{yang2022learning}   & 66.3 & 68.7 & 75.7 & 83.6  \\ 
CAL~\cite{gu2022clothes}             & 66.9 & 69.6 & 79.1 & 85.0  \\ 

MPANet~\cite{wu2021discover}& 68.2 & 70.6 & 80.7 & 82.8 \\
MMN~\cite{zhang2021towards}  & 66.9 & 70.6 & 84.1 & 91.6 \\
DCLNet~\cite{sun2022not}  & 65.3 & 70.8 & 74.3 & 81.2 \\
MAUM~\cite{liu2022learning} & 68.8 & 71.7 & 85.1 & 87.9 \\
DEEN~\cite{Zhang_Wang}      & 71.8 & 74.7 & 85.1 & 91.1 \\

\midrule
ChatReID                & \textbf{83.6} & \textbf{86.8} & \textbf{95.8} & \textbf{96.5}  \\ 
 \bottomrule
\end{tabular}}
\vspace{-0.1in}
\end{table}

\begin{table*}[t]
\small
\centering
\caption{Ablation Study. Performance comparison of different stage combinations in our hierarchical progressive learning strategy across ten benchmarks. The abbreviation `MM01` refers to the SYSU-MM01 dataset, `CUHK` refers to the CUHK-PEDES dataset, and `ICFG` refers to the ICFG-PEDES dataset.}
\vspace{-0.1in}
\label{aba}
   \setlength{\tabcolsep}{1.7mm}{
\begin{tabular}{lccccccccccc}
\toprule
\multicolumn{1}{c}{\multirow{2}{*}{\textsc{Methods}}} & \multicolumn{3}{c}{\textbf{Standard ReID}} & \multicolumn{2}{c}{\textbf{CC-ReID}} & \multicolumn{2}{c}{\textbf{VI-ReID}} & \multicolumn{3}{c}{\textbf{T2I-ReID}} &  \multirow{2}{*}{\textbf{Avg.}} \\ 
\cmidrule(r){2-4} \cmidrule(r){5-6} \cmidrule(r){7-8} \cmidrule(r){9-11} 

                         & Market1501 & MSMT17 & CUHK03 & PRCC & LTCC & MM01 & RegDB & CUHK & ICFG & RSTPReid  \\ \hline
Stage 3                   &59.2  &41.5& 60.3 &  31.2 & 35.4  & 11.4 & 12.7 &  65.1 & 49.9 & 43.5 &41.0 \\ 
Stage 2 + Stage 3          & 92.1 & 79.5 & 85.7 & 60.2 & 55.1 & 76.2 & 74.7 & 77.8 & 62.5 & 71.4 & 73.5  \\ 
 \midrule
ChatReID                  & \textbf{96.4} & \textbf{87.5}  & \textbf{91.3} & \textbf{76.3} & \textbf{75.6} & \textbf{83.6} & \textbf{95.8} & \textbf{80.1} & \textbf{70.5} & \textbf{73.1} & \textbf{83.0} \\ \bottomrule
\end{tabular}}
\vspace{-0.1in}
\end{table*}

\begin{table}[t]
    \centering
    \small
    \caption{Comparison of ChatReID with SOTA methods across three T2I Re-ID datasets.}
    \vspace{-0.1in}
    \label{t2i}
    \setlength{\tabcolsep}{1.25mm}{
    \begin{tabular}{lcccccc}
        \toprule
        \multicolumn{1}{c}{\multirow{2}[0]{*}{\textsc{Methods}}} & \multicolumn{2}{c}{\textbf{CUHK-PEDES}} & \multicolumn{2}{c}{\textbf{ICFG-PEDES}} & \multicolumn{2}{c}{\textbf{RSTPReid}} \\
        \cmidrule(r){2-3} \cmidrule(r){4-5} \cmidrule(r){6-7}
                 & mAP & Rank-1 & mAP & Rank-1 & mAP & Rank-1 \\
        \midrule
        IRRA~\cite{jiang2023cross} & 66.1 & 73.4 & 38.1 & 63.5 &- & 60.2\\
        RDE~\cite{qin2024noisy}  & 67.6 & 75.9 & 40.1 & 67.7 &50.9 & 65.4\\
        WoRA~\cite{sun2024data}  & 67.2 & 76.4 & 42.6 & 87.5 & 52.5 & 66.9 \\
        RaSa~\cite{bai2023rasa}        & 69.4 & 76.5 & 41.3 & 65.3 & 52.3 & 67.0 \\
        APTM~\cite{yang2023towards}   & 66.9 & 76.5 & 41.2 & 68.5 & - & 67.5 \\
        MARS~\cite{ergasti2024mars}   & 71.7 & 77.6 & 44.9 & 67.6 & 52.9 & 67.6\\
        IRM~\cite{he2024instruct}  &66.5 &74.2      & - & - & - & -\\
        \midrule
        ChatReID    & \textbf{80.1} & \textbf{83.8} & \textbf{70.5} & \textbf{72.9} & \textbf{73.1} & \textbf{75.0} \\
        \bottomrule
    \end{tabular}}
    \vspace{-0.1in}
\end{table}
\noindent \textbf{Visible-infrared person ReID (VI-ReID).}
As shown in Tab.~\ref{vi}, ChatReID also demonstrates significant improvements on VI-ReID datasets. Specifically, it achieves an 11.8\% mAP increase on SYSU-MM01 and a 10.7\% mAP increase on RegDB. These gains are primarily attributed to our approach in Stage 2, where we treated the VI-ReID task as an image-image matching task and slightly increased the proportion of VI data in the tuning process.

\noindent \textbf{Text-to-Image Person Re-ID.}
As shown in Tab.~\ref{t2i}, ChatReID achieves remarkable gains in text-to-image person Re-ID, especially on the ICFG-PEDES and RSTPReid datasets, where previous performance was relatively low. ChatReID reaches an mAP of 70.5\% on ICFG-PEDES, marking a 25.6\% improvement, and an mAP of 73.1\% on RSTPReid, a 20.2\% increase. 
These gains are attributed to two factors: existing VLMs have a baseline text-to-image matching capability, and our image-text pair dataset structure enhances person-specific matching across all training stages.

\noindent \textbf{Attribute-Based Person Retrieval.}
To the best of our knowledge, there is currently no straightforward and effective method for the Attribute-Based Person Retrieval task. Existing text-to-image person Re-ID methods primarily focus on retrieving images based on complete textual descriptions rather than specific attributes. To address this gap, we introduced attribute-based retrieval as a training objective in Stage 3.
To assess the effectiveness of our approach, we reorganized the dataset by randomly sampling $N$ images from an attribute-annotated dataset. Our goal was to retrieve images that matched specified attributes, such as clothing or accessories.
Experimental results indicate that ChatReID can efficiently retrieve images based on attribute criteria, highlighting its practical value for real-world applications. Additional experimental results are presented in the supplementary material.

\noindent \textbf{Adaptive Multi-Modal Person Retrieval.}
 Considering practical application requirements, we design a training task for Adaptive Multi-Modal Person Retrieval. To evaluate the model's performance on this task, we conducted experiments that included additional textual information as auxiliary input for image-image retrieval tasks.
The experimental results demonstrate that ChatReID can efficiently handle person retrieval with multi-modal inputs. Detailed results and analyses are provided in the supplementary material.

\subsection{Ablation Study \label{sec:ablation}}
\textbf{Three-Stage HPT Strategy.} To validate the effectiveness of our three-stage HPT strategy, we compare experimental results across different stage combinations, as shown in Tab.~\ref{aba}. Initially, we report the results from directly applying Stage 3 training. While Stage 3 achieves baseline performance, the results are suboptimal, indicating that current VLMs have limited inherent capabilities for image-to-image and image-text pedestrian matching and require specialized training for optimal performance.
Our experiments further reveal that combining only Stage 1 and Stage 3 made it difficult for the model to converge, and thus, those results were not reported. However, incorporating Stage 2 into the training process significantly improved the model’s performance, achieving an average mAP of 73.5\%. The fine-grained image retrieval tuning in Stage 2 led to substantial improvements, providing strong evidence for the effectiveness of our approach.
Finally, when Stage 1 training is also included, the model's performance is further enhanced, reaching state-of-the-art (SOTA) results across ten benchmarks. This final improvement underscores the overall effectiveness of our hierarchical three-stage tuning strategy.

\noindent \textbf{VLMs Model Size and Performance Trade-off.} In our validation experiments, we evaluated VLMs of different sizes under each setting. The results are shown in Tab.~\ref{tab:vlm}. As the model size increases, accuracy improves. Both the 2B and 7B models achieve comparable performance. This can be attributed to the fact that the Re-ID task is relatively less complex compared to VLM training scenarios, allowing the 2B model to effectively capture the necessary representations. Given that the 7B model incurs substantially higher computational costs without yielding significant performance gains, we chose the 2B model.

\begin{table}[t]
    \centering
    \small
    \caption{Performance on VLMs with different model size. The data in the table represents Rank-1.}
    \vspace{-0.1in}
    \label{tab:vlm}
   \setlength{\tabcolsep}{1mm}{
    \begin{tabular}{lcccc}
        \toprule
        \textsc{\#Parameter} & \textbf{MSMT17} & \textbf{LTCC} & \textbf{RegDB} & \textbf{CUHK-PEDES}\\
 
        \midrule
        0.5 B & 38.1 & 27.4 & 4.3 & 55.1 \\
        2 B  & 90.1 & \textbf{82.7} & \textbf{96.5} & 83.8 \\
        7 B  & \textbf{91.3} & 82.5 & 96.3 & \textbf{87.7}  \\
        \bottomrule
    \end{tabular}}
\vspace{-0.1in}
\end{table}

\section{Broad Impact and Conclusion}
\noindent \textbf{Broad Impact.} We utilize publicly available, high-quality academic datasets for our research. These datasets were curated by various institutions across diverse scenarios, which helps mitigate potential biases in the images. It is important to note that, due to privacy policies, we do not use the DukeMTMC~\cite{ristani2016performance} dataset. We will release the code, datasets, and models upon acceptance to facilitate further research. To ensure responsible use, we will adhere to a strict application protocol to prevent our work from being used for any unethical or illegal purposes.

\noindent \textbf{Conclusion.} This paper introduces ChatReID, a novel framework for person re-identification (Re-ID) that leverages the advanced capabilities of vision-language models (VLMs). Our ChatReID is a text-side-dominated framework for person Re-ID. Different from traditional methods that calculates image feature distance as similarity, ChatReID interprets the task requirements from the input text description and performs similarity inference on the given person images accordingly.
By implementing a Hierarchical Progressive Tuning (HPT) strategy, we progressively enhance the model’s ability to achieve fine-grained, identity-level retrieval.
The system’s VQA-based inference format simplifies complex Re-ID tasks, making it more accessible to non-experts, while its flexibility allows dynamic input combinations and adjustments. Our work lays a strong foundation for future advancements in person Re-ID, showcasing the potential of LVLMs in practical applications.

{
    \small
    \bibliographystyle{ieeenat_fullname}
    \bibliography{main}
}


\end{document}